\documentclass{article}

\usepackage[preprint]{neurips_2025}

\usepackage[utf8]{inputenc} % allow utf-8 input
\usepackage[T1]{fontenc}    % use 8-bit T1 fonts
\usepackage{hyperref}       % hyperlinks
\usepackage{url}            % simple URL typesetting
\usepackage{booktabs}       % professional-quality tables
\usepackage{amsfonts}       % blackboard math symbols
\usepackage{nicefrac}       % compact symbols for 1/2, etc.
\usepackage{microtype}      % microtypography
\usepackage{xcolor}         % colors
\usepackage{multirow}
\usepackage{booktabs}
\usepackage{graphicx} % for \resizebox
\usepackage{amsmath}  % for \pm

\usepackage{amsmath,amsfonts,bm}

\def\eqref#1{(\ref{#1})}
\def\1{\bm{1}}

\DeclareMathAlphabet{\mathsfit}{\encodingdefault}{\sfdefault}{m}{sl}
\SetMathAlphabet{\mathsfit}{bold}{\encodingdefault}{\sfdefault}{bx}{n}

\newcommand{\E}{\mathbb{E}}

\newcommand{\Var}{\mathrm{Var}}

\usepackage[percent]{overpic} 

\usepackage{arydshln}

\usepackage{algorithm}
\usepackage{algpseudocode}
\usepackage{amsmath}

\usepackage{amsmath}
\usepackage{amssymb}
\usepackage{mathtools}
\usepackage{amsthm}
\usepackage[capitalize,noabbrev]{cleveref}

\theoremstyle{plain}

\theoremstyle{definition}

\theoremstyle{remark}

\usepackage[textsize=tiny]{todonotes}

\title{Path Gradients after Flow Matching}

\author{%
   Lorenz Vaitl \\
  \texttt{lorenz.vaitl@outlook.com} \\
  \And
    Leon Klein\\ 
    Freie Universität Berlin\\
    \texttt{leon.klein@fu-berlin.de} \\
}

\begin{document}

\maketitle

\begin{abstract}
Boltzmann Generators have emerged as a promising machine learning tool for generating samples from equilibrium distributions of molecular systems using Normalizing Flows and importance weighting.
Recently, Flow Matching has helped speed up Continuous Normalizing Flows (CNFs), scale them to more complex molecular systems, and minimize the length of the flow integration trajectories.
 We investigate the benefits of using Path Gradients to fine-tune CNFs  initially trained by Flow Matching, in a setting where the target energy is known. 
 Our experiments show that this hybrid approach yields up to a threefold increase in sampling efficiency for molecular systems,
 all while using the same model, a similar computational budget and without the need for additional sampling.
 Furthermore, by measuring the length of the flow trajectories 
 during fine-tuning, we show that Path Gradients largely preserve the learned structure of the flow.
 \end{abstract}

\section{Introduction}
\label{submission}

Generative models, ranging from GANs \citep{goodfellow2014generative} and VAEs \citep{kingma2013auto} to Normalizing Flows \citep{RezendeEtAl_NormalizingFlows, papamakarios19_normal_flows_probab_model_infer} and Diffusion Models \citep{ho2020denoising, song2020score}, have advanced rapidly in recent years, driving progress both in media generation and in scientific applications such as simulation-based inference. While scientific workflows often incorporate domain-specific symmetries, they tend to under-exploit a crucial resource: the unnormalized target density.

% No need for additional energy evaluations
Boltzmann Generators \citep{noe2019boltzmann} are typically trained either via self-sampling \citep{boyda2021sampling, nicoli2020asymptotically, invernizzi2022skipping,midgley2022flow}, leveraging gradients from the target distribution, or using samples from the target without incorporating gradient information \citep{nicoli2023detecting,klein2023equivariant,pmlr-v238-draxler24a, klein2024transferable}.
However, these approaches each ignore complementary parts of the training signal: either the data or its local geometry.
Notably, first-order information evaluated at target samples remains underused, despite its potential to improve training. 
In this work, we close this gap by fine-tuning Continuous Normalizing Flows, initially trained with Flow Matching, using Path Gradients \citep{roeder2017sticking, Vaitl2024} on samples from the target distribution. 
Furthermore, our approach requires computing target gradients only once per training sample, avoiding the potentially high cost of repeatedly computing gradients on newly generated samples for self-sampling.

Flow Matching, a method for training CNFs, is based on target samples. It is closely related to diffusion model training and has recently gained traction for its simulation-free training and strong empirical performance, both in generative modeling benchmarks \citep{lipman2022flow,esser2024scaling, jin2024pyramidal} and in scientific domains \citep{stark2024harmonic, jing2024alphafold, klein2024transferable}.
Here, we investigate how incorporating Path Gradients on samples from the target distribution, enhances CNF performance post flow-matching.

Path gradients are low variance gradient estimators, which have strong theoretical guarantees close to the optimum \citep{roeder2017sticking} and incorporate gradient information from both, the variational and the target distribution \citep{vaitl2023fast}.
While  they have been adopted for training Normalizing Flows in the field of Lattice Gauge Theory \citep{bacchio2023learning, kanwar2024flow, abbott2305normalizing} and also variational inference \citep{agrawal2024disentangling, andrade2023loft}, they remain underused in Biochemistry.
In this work, we explore the potential of Path Gradient fine-tuning for Boltzmann Generators with promising results.
This indicates that even though training with Path Gradients is orders of magnitude slower \textit{per iteration}, its use of additional information and low variance allows it to outperform Flow Matching \textit{within the same computational constraints} in fine-tuning.
%@ Leon are these results SOTA? yes

We make the following main contributions:
\begin{itemize}
    \item We propose hybrid training using Flow Matching for pre-training and Path Gradients for fine-tuning. We do not require additional samples beyond the original force-labeled data, which were required already for generating the data. In order to ensure fast training, we are the first to optimize CNFs using Path Gradients for the forward KL and making use of the augmented adjoint method \citep{vaitl2022path}. % and Path Gradients \citep{bauer2021generalized, vaitl2023fast}.
    \item We show for many particle systems as well as alanine dipeptide that this fine-tuning approach can triple importance-sampling efficiency within the same computational budget.
    \item We investigate Path Gradients' impact on properties trained during Flow Matching, namely flow trajectory length and the MSE Flow Matching loss, and show that these are mostly unaffected by the fine-tuning. 
\end{itemize}

\begin{figure}[h]
\begin{center}
\includegraphics[width=\textwidth]{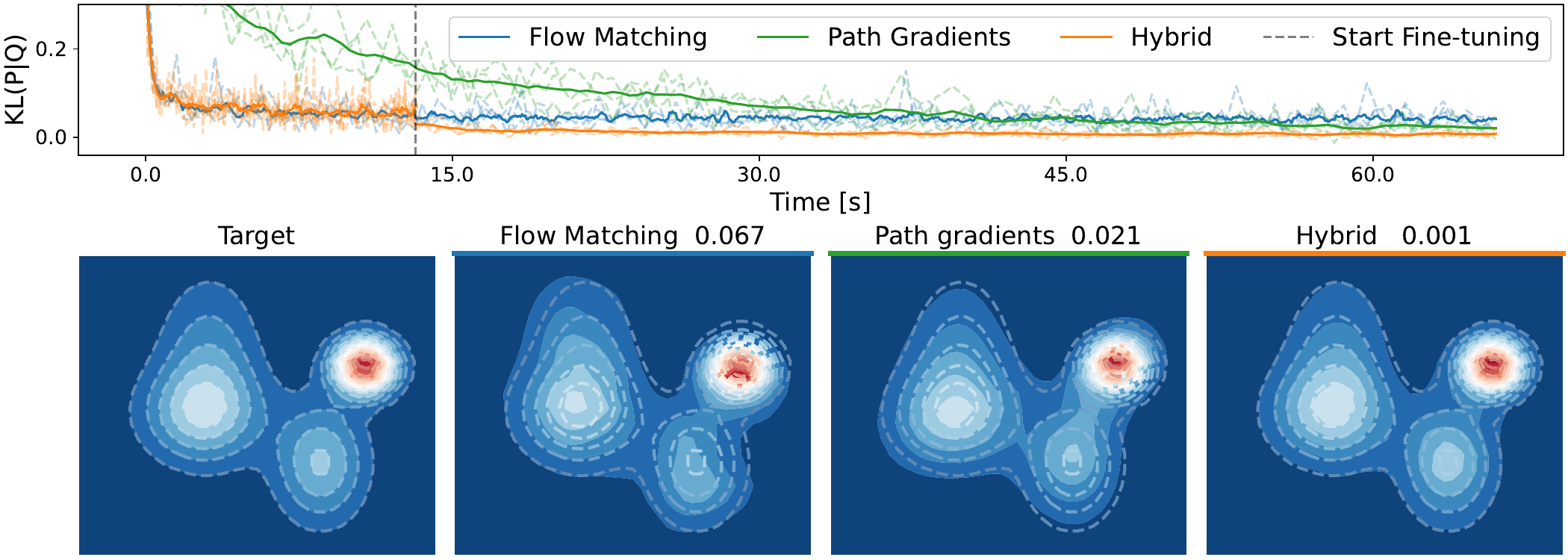}
\caption{Training a CNF on a simple 2D Gaussian Mixture Model.
Comparison between pure training with Flow Matching, pre-training with Flow Matching and fine-tuning with Path Gradients and pure training with Path Gradients.
We see that given the same wall-time hybrid training performs best in terms of forward KL divergence.
The bottom row shows the target and the final model after training.}
\label{fig:2d-GMM}
\end{center}
\end{figure}

\section{Method}
Generative models have recently garnered significant interest for biochemical applications \citep{noe2019boltzmann, jumper2021highly, abramson2024accurate}. In this work, we focus on the application of enhancing or replacing molecular dynamics simulations. 
%The aim of a large branch of AI4Science lies in the field of Boltzmann Generators.
%Amortized inference “pays” the upfront cost of model training in exchange for dramatically accelerated sampling, effectively speeding up simulations. 

\subsection{Boltzmann Generators}
Boltzmann Generators \citep{noe2019boltzmann} aim to asymptotically generate unbiased samples from the equilibrium Boltzmann distribution 
\begin{equation}
    p(x) = \frac{1}{Z} \exp(-U(x))
\end{equation}
with an energy $U(x)$ and an unknown normalization factor $Z = \int \exp(-U(x)) d x$. The Boltzmann distribution describes the probability of states in a physical system at thermal equilibrium. Traditional sampling methods, such as Markov Chain Monte Carlo (MCMC) and Molecular Dynamics (MD) simulations, generate samples sequentially. In contrast, Boltzmann Generators are designed to produce independent samples directly from the target distribution. There are many different instances of Boltzmann Generators, many focus on molecular systems \citep{dibak2021temperature, kohler2021smooth, midgley2022flow, ding2021deepbar, ding2021computing, kim2024scalableNormFlows, rizzi2023multimap, tamagnone2024coarse, schonle2024sampling, klein2024transferable, tan2025scalable}, while others are trained to sample lattice and other many particle systems \citep{wirnsberger2020targeted, Ahmad2022, nicoli2020asymptotically, nicoli2021estimation, schebek2024efficient, abbott2305normalizing, kanwar2024flow}.

Boltzmann Generators employ a Normalizing Flow to map samples from a simple base distribution into a learned sampling distribution $q_\theta$, which is trained to approximate the target Boltzmann distribution.
We can use Boltzmann Generators to obtain asymptotically unbiased estimators for observables over $p(x)$ using the sampling distribution $q_\theta$ and (self-normalized) importance sampling \citep{noe2019boltzmann, nicoli2020asymptotically},
\begin{equation}\label{eq:importance}
    \E_{p(x)} \left[ \mathcal O(x) \right] =  \E_{q_\theta(x)} \left[ \frac{p(x)}{q_\theta(x)} \mathcal O(x) \right] \, ,
\end{equation}
for an observable $\mathcal O: \mathbb R^d \rightarrow \mathbb R $.
For self-normalized importance sampling, we still have to estimate the normalization constant
\begin{equation}
    Z = \int \exp(-U(x)) d x = \E_{q_\theta(x)} \left[ \frac{p(x)}{q_\theta(x)}\right]
\end{equation}
 with an MC estimator.
This separates Boltzmann Generators from related generative methods, which only generate approximate samples from the Boltzmann distribution \citep{jing2022torsional, abdin2023pepflow, klein2023timewarp, schreiner2023implicit, jing2024alphafold, lewis2024scalable}.

We can estimate the efficiency of importance sampling using the (relative) effective sampling size (ESS). The ESS 
\begin{equation}
    \text{ESS}:=\left({\E_{q_\theta}\left[\left(\frac{p}{q_\theta} (x)\right)^2\right]}\right)^{-1} = \left({\E_{p}\left[\frac{p}{q_\theta}(x)\right]}\right)^{-1} \in [0,1]
\end{equation}
can be estimated either on samples from the model $q_\theta$ or the target distribution $p$.
Roughly speaking, the ESS tells us how efficiently we are able to sample from the target distribution. 
The estimator %-- the reverse ESS -- 
ESS$_q$, based on samples from $q_\theta$, might fail to detect missing modes in the target distribution, while %-- the forward ESS --  
ESS$_p$ requires samples from the target distribution.
Both estimators exhibit high variance and might overestimate the performance if too few samples are being used.
For more information confer \citep{nicoli2023detecting}.

\subsection{Continuous Normalizing Flows}

Neural ODEs, introduced by \citet{chen2018neural}, parameterize a continuous-time transformation through a neural-network-defined vector field $v_{\theta}$. By employing the adjoint sensitivity method \citep{pontryagin1987mathematical}, gradients are computed in a backpropagation-like fashion with constant memory.
By computing the ODE of the divergence, we can compute the change in probability, which lets us use a Neural ODE as a Normalizing Flow \citep{chen2018neural, grathwohl2018ffjord}.

Using a simple base distribution $q_0(x_0)$ and the transformation 
\begin{equation}
    T_\theta(x_0) = x_0 + \int_0^1 v_\theta(x_t, t) d t  \, , \label{eq:NODE}
\end{equation}
we obtain a distribution 
$q_{\theta}$, which we can sample from and compute the probability
\begin{align}
    \log q_{\theta}(T_\theta(x_0)) &= \log q_0(x_0) - \log \det \left(  \frac{\partial T_\theta(x_0)}{\partial x_0} \right) \nonumber \\
    &= \log q_0(x_0) - \int_0^1 \text{Tr}\left( \frac{\partial v_\theta(x_t,t)}{\partial x_t} \right) d t  \, . \label{eq:Divergence}
\end{align}

For training samples from the target distribution, a straightforward way of approximating the target density is by maximum likelihood training, i.e. by minimizing
\begin{align}
    \mathcal L_{\text{ML}}(\theta) = -\E_{p(x)} \left[ \log q_\theta(x) \right] \, . \label{eq:ML}
\end{align}
Minimizing \cref{eq:ML} is equivalent to minimizing the forward KL divergence up to a constant
\begin{align}
    KL(p|q_\theta) = \E_{p(x)} \left[\log p(x) - \log q_\theta(x) \right] \overset  c =\mathcal L_{\text{ML}}(\theta)\, . \label{eq:InitialKLForward}
\end{align}
As \cite{kohler2020equivariant} showed, we can construct equivariant CNFs by using an equivariant network as the vector field.  

\subsection{Flow Matching}
\label{sec:Flow-Matching}
Inspired by advances in diffusion models, Flow Matching \citep{lipman2022flow,  albergo2023stochastic, liu2022flow} has emerged as a promising competitor to diffusion models \citep{esser2024scaling}.
Flow Matching enables us to train CNFs in a simulation free manner, i.e. training without the need of the expensive integration of the ODEs \eqref{eq:NODE} and \eqref{eq:Divergence}.

The idea behind Flow Matching is to posit a base probability $p_0$ and a vector field $u_t(x)$, which generates $p_t$ from the base density $p_0$, such that the final probability $p_1$ equals the target density $p$.
Given these components, Flow Matching aims to minimize the mean squared error between the target vector field $u_t(x)$ and the learned vector field $ v_{\theta, t}(x)$
\begin{align}
    \mathcal L_{\text{FM}}&(\theta) = \E_{t \sim \mathcal U(0,1), x \sim p_t(x)} \left[ || v_{\theta}(x, t) - u_t(x) ||^2\right] \, . \label{eq:FullFlowMatchingObj}
\end{align}
since $p_t$ and $u_t$ are not known, it is intractable to compute the objective. Nevertheless, \citet{lipman2022flow} showed that we can construct a conditional probability path $p(x|z)$ and corresponding vector field $u_t(\cdot| z)$ conditioned on $z = (x_0, x_1)$, which has identical gradients to Eq.~\eqref{eq:FullFlowMatchingObj}.
This yields the conditional Flow Matching objective 
\begin{align}
    \mathcal L_{\text{CFM}}&(\theta) = 
    \E_{t \sim \mathcal U(0,1), x \sim p_t(x|z)} \left[|| v_\theta(x, t) - u_t(x|z) ||^2 \right] \, . \label{eq:CFM}
\end{align}

In this framework, there are different ways of constructing the conditional probability path. We here focus on the parametrization introduced in \citet{tong2023conditional}, which results in optimal integration paths
\begin{align}
z = (x_0, x_1) \quad &\textrm{and}\quad p(z) = \pi(x_0,x_1) \\
u_t(x | z) = x_1 - x_0 \quad &\textrm{and}\quad p_t(x | z) = \mathcal{N}(x | t \cdot x_1 + (1 - t) \cdot x_0, \sigma^2),
\end{align}
where $\pi(x_0,x_1)$ denotes the optimal transport map between $p_0$ and $p_1$. As this map is again intractable, it is only evaluated per batch during training. We refer to this parametrization as \textit{OT FM}. In \citet{klein2023equivariant,song2023equivariant} the authors show that we can extend this to highly symmetric systems, such as many particle systems and molecules, and still obtain optimal transport paths. The approach enforces the system’s symmetries in the OT map by computing the Wasserstein distance over optimally aligned samples with respect to those symmetries. Hereafter, we denote this parameterization as \textit{EQ OT FM}. For more details on optimal transport Flow Matching refer to \citet{tong2023conditional} and \citet{klein2023equivariant,song2023equivariant}.
Note, that we obtain the original formulation in \citet{lipman2022flow} by using $p(z)=p_0(x_0)p_1(x_1)$, we will refer to it as \textit{standard FM}.

%$p_0 = \mathcal N(0,1)$ and 
%\begin{align}
%    p_t(x|x_1) &= \mathcal N(x| \mu_t(x_1), \sigma_t(x_1)) \, ,
%\end{align}
%with $\mu_t = t x_1$ and $\sigma_t = 1 - (1 - \sigma_{\text{min}}) t$, for some $\sigma_{\text{min}} > 0$.
\subsection{Path Gradients}
\label{sec:Path-Grads}

Introduced in the context of variational auto-encoders \citep{roeder2017sticking, tucker2018doubly}, Path Gradient estimators have improved performance, especially when applied to normalizing flows \citep{agrawal2020advances, vaitl2022gradients, vaitl2022path,  agrawal2024disentangling}.
In the field of simulating Lattice Field Theories, they have become a go-to tool for training flows \citep{bacchio2023learning, kanwar2024flow, abbott2305normalizing}.
In these applications, the reverse Kullback-Leibler divergence $KL(q_\theta|p)$ is minimized by self-sampling, i.e. without existing samples from the target density, but only on samples from $q_\theta$ generated while training.\\
Path gradients are unbiased gradient estimators which have low variance close to the optimum.
We obtain Path Gradients by separating the total derivative of a reparametrized gradient  for the reverse KL
\begin{align}
    \frac{d}{d \theta} KL(q_\theta|p) = \E_{q_0(x_0)} \left[\frac{d}{d \theta} \log \frac{q_\theta(T_\theta(x_0))}{p(T_\theta(x_0))} \right]
\end{align}
 into two partial derivatives
\begin{align}
    \E_{q_0(x_0)} \Bigg[&\underbrace{\frac{\partial }{\partial x_1} \log \frac{q_\theta(x_1)}{p(x_1)} \cdot \frac{\partial T_{\theta}(x_0)}{\partial \theta} }_{\text{Path Gradient of } KL(q_\theta|p)} + \underbrace{\frac{\partial \log q_\theta(x_1)}{\partial \theta}\Big|_{x_1 = T_\theta(x_0)}}_{\text{Score term}} \Bigg] \, ,
\end{align}
called the Path Gradient and the score term~\footnote{The score term $\frac{\partial \log q_\theta(x)}{\partial \theta}$ is not to be confused with the force term $\frac{\partial \log q_\theta(x)}{\partial x}$, which is often called the score in the context of diffusion models \citep{song2020score}.}.

\citet{roeder2017sticking} observed, that the martingale score term
vanishes in expectation, but has a non-zero variance. \citet{vaitl2022gradients} showed that its variance is $\frac 1 N I_\theta$ for a batch of size $N$ and the Fisher Information  $I_\theta$ of the variational distribution $q_\theta$.
The low variance close to the optimum allows Path Gradient estimators to "stick-the-landing" \citep{roeder2017sticking}, i.e. having zero variance at the optimum, making it an ideal tool for fine-tuning.
\citet{vaitl2023fast} showed that Path Gradient estimators incorporate additional information about the derivative of both target and variational distribution, opposed to the reparametrized gradient gradient estimator \citep{kingma2013auto, mohamed2019monte} (see \cref{app:estimators} for a detailed explanation).
They hypothesized that this leads to less overfitting and a generally better fit to the target density (see \citet{Vaitl2024}).
These theoretical aspects make Path Gradients a promising suitor for fine-tuning Boltzmann Generators.

\paragraph{Path Gradients on given samples}
Recently, Path Gradient estimators have been proposed for samples from distributions different from the variational density $q_\theta$ \citep{bauer2021generalized}.
By viewing the forward KL$(p_1|q_\theta)$ at $t=1$ as a reverse KL$(p_{0, \theta}| q_0)$ at $t=0$, we can straightforwardly apply Path Gradients \citep{vaitl2023fast}.
Here, we assume $p_{0, \theta}$ to be a (Continuous) Normalizing Flow from $p_1$ as its base density and $T^{-1}_\theta$ to be the parametrized diffeomorphism.

\begin{align}
    KL(p_1|q_\theta) &= E_{p_1(x_1)}[ \log p_1(x_1) - \log q_\theta(x_1)] \nonumber \\
    &= E_{p_1(x_1)}\Big[ \log p_1(x_1) -  
      \underbrace{\left(\log q_0(T^{-1}_\theta(x_1)) - \log \det \left| \frac{\partial T^{-1}_\theta(x_1)}{\partial x_1} \right) \right| }_{= \log q_\theta(x_1)} \Big] \\
    &= E_{p_1(x_1)}\Big[  
     \underbrace{\left(\log p_1(x_1) - \log \det \left| \frac{\partial T^{-1}_\theta(x_1)}{\partial x_1} \right| \right)}_{= \log p_{0, \theta}(x_0)}   -\log q_0(T^{-1}_\theta(x_1))\Big] \nonumber \\
    &= E_{p_{0,\theta}(x_0)}\Big[  
     \log p_{0, \theta}(x_0)   -\log q_0(x_0)\Big]  = KL(p_{0, \theta}|q_0) \, . \label{eq:forwardKL}
\end{align}
This different view lets us compute Path Gradients with the same ease as for the reverse KL, but on given samples from the target distribution.

While \citet{vaitl2022gradients} and the present work both use Path Gradients, both minimize different losses on different samples. Our work optimizes the Forward KL$(p_1|q_\theta)$ with path gradients on existing samples from $p$, as proposed in \citet{vaitl2023fast} using the algorithm proposed in \citet{vaitl2022gradients}. 
\citet{vaitl2022gradients} minimize the reverse KL$(q_\theta|p_1)$ via self-sampling, i.e. on samples from the model $q_\theta$.
Training via self-sampling has many drawbacks, mainly: modes of the target $p$ can be entirely missed, which invalidates all asymptotic guarantees and breaks importance sampling, see e.g. \citet{nicoli2023detecting}.
This becomes increasingly likely in higher dimensions.
Further, the forces for the newly generated samples have to be evaluated, which can increase the cost of training or lead to unstable optimization behavior.

Since the publication \citet{vaitl2022path}, Flow Matching has been established as the de facto training method for CNFs. Our work investigates and combines the performance of Flow Matching and Path Gradients. Specifically, we investigate the effect of Path Gradients on the inner workings of the CNF, like e.g. trajectory length, while \citet{vaitl2022path} simply looked at the performance compared to standard self-sampling losses.

In \cref{app:Theory}, we zoom in on the properties of the different estimators.
We show that while Path Gradients exhibit zero variance at the optimum, generically, Flow Matching does not.
We further explicate the theoretical properties of the existing estimators for training via Maximum Likelihood, Path Gradients and Flow Matching.

While the naive calculation of Path Gradients requires significantly more compute and memory than Flow Matching, 
we take several steps to obtain constant memory and speed ups in the next section.
%, using the augmented adjoint, as proposed in \citet{vaitl2022path}, only requires roughly a third more compute and the same memory requirements.
As we show experimentally,
%while this is still significantly more compute than Flow Matching,
fine-tuning with Path Gradient for a few epochs significantly improves the performance for Continuous Normalizing Flows compared to only using Flow Matching.

\section{Path Gradients and Flow Matching}
\label{sec:PG-vs-FM}

\begin{figure}
    \centering
    \includegraphics[width=1\linewidth]{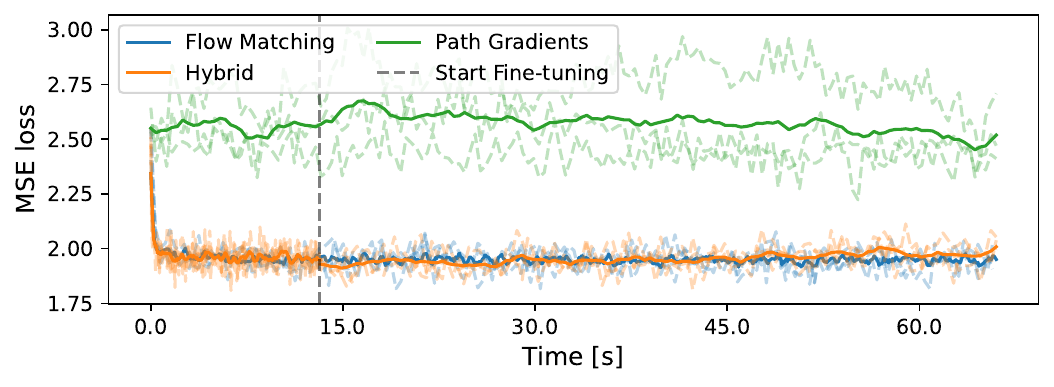}
    \caption{FM loss objective \ref{eq:CFM} during training on 2D GMM averaged on three runs. Training with Path gradients leaves the MSE largely unchanged.}
    \label{fig:MSE-loss-GMM}
\end{figure}

To build an intuition for the behavior of Flow Matching (FM) versus Path Gradient (PG) estimators, we start with a simple 2D Gaussian Mixture Model.
We compare three training strategies: 1) FM only, 2) PG only, and 3) a hybrid approach, using FM to quickly approximate the target distribution $p_1$, and subsequently applying PG for fine-tuning.
Although slower per training step, PG has strong theoretical guarantees near the optimum.\\
For Flow Matching, we use the standard formulation proposed in \citet{lipman2022flow}.
The dynamics $v_\theta$ is modeled using a four-layer fully connected neural network with 64 units per layer and ELU activations. Given the simplicity of the model and task, we ignore memory usage in this setup.
All experiments are run on a CPU and complete in roughly one minute.\\
As shown in \cref{fig:2d-GMM}, FM training rapidly improves initially but soon reaches a plateau with slow improvements after.
PG training in contrast, progresses more slowly but eventually reaches and surpasses the FM plateau.
The hybrid strategy, i.e. beginning with FM and switching to PG, results in the best performance, combining fast convergence with improved final accuracy. \\
Interestingly, as shown in \cref{fig:MSE-loss-GMM}, applying PG after FM has little effect on the MSE loss defined in \cref{eq:CFM}, suggesting that PG fine-tunes the variational distribution $q_\theta$ without significantly altering the dynamics $v_\theta$.
Nonetheless, this subtle adjustment leads to visibly better samples and improved density matching (cf. \cref{fig:2d-GMM}).
%Lastly, \cref{fig:KLparts-2dGMM} shows the KL$(p|q_\theta)$ parts. 
%
\\
In the 2D case, we compute the divergence term in \cref{eq:Divergence} exactly by directly evaluating the trace of the Jacobian. 
However, this approach scales quadratically with the number of dimensions and quickly becomes computationally prohibitive in higher-dimensional settings. 
A practical solution is Hutchinson's trace estimator \citep{hutchinson1989stochastic}, which provides us faster but noisy estimators for the divergence. \\
For ODE integration, we use a fourth-order Runge–Kutta (RK4) scheme with 15 time steps, resulting in 60 function evaluations per integration trajectory. Under these settings, training on a single batch is empirically about 275 times faster with Flow Matching compared to Path Gradients. 
This discrepancy is expected, as the cost of function evaluations differs substantially between the two methods.

While these results are promising, the 2D setup allows exact computation of the divergence and may not generalize to higher-dimensional or more complex tasks. Moreover, memory usage becomes a significant bottleneck in these more realistic scenarios.

\paragraph{Augmented Adjoint Method}
In order to have constant memory irrespective of the number of function evaluations, we adapt the augmented adjoint method \citep{vaitl2022path} for Path Gradient estimators for the forward KL divergence \cref{eq:forwardKL}
\begin{align}
    \frac{d}{d \theta} KL(p_{0, \theta}| q_0) &= \E_{x_1 \sim p_1} \left[\frac{\partial }{\partial x_0 } \left(\log \frac{p_{0, \theta}}{q_0}(x_0) \right) \frac{\partial T_\theta^{-1}(x_1)}{\partial \theta} \right] \, . \label{eq:PG-FWKL}
\end{align}

In order to compute the force of the variational distribution
\begin{align}
    \frac{\partial \log p_{0, \theta}(x_0)}{\partial x_0}
\end{align}
we use the augmented adjoint method by solving the ODE
\begin{align}
    \frac{d}{d t} \frac{\partial \log p_{t, \theta}(x_t)}{\partial x_t} = - \frac{\partial \log p_{t, \theta}(x_t)}{\partial x_t}^T \frac{\partial v_\theta(x_t, t)}{\partial x_t} - \frac{\partial }{\partial x_t} \text{Tr}\left(\frac{\partial v_\theta(x_t,t)}{\partial x_t} \right) \label{eq:augAdjoint}
\end{align}
with initial condition
\begin{align}
    \frac{\partial \log p_{1, \theta}}{\partial x_1} = \frac{\partial \log p_1(x_1)}{\partial x_1} \, 
\end{align}
from $t=1$ to $0$
\citep{vaitl2022path}.
%\todo{Assumptions: Picard-Lindeloef...}

Computing Path Gradients requires computing the gradient of the divergence 
\begin{align}
 \text{Tr}\left(\frac{\partial v_\theta(x_t,t)}{\partial x_t} \right) \, 
\end{align}
as well as the ODEs \ref{eq:NODE} and \ref{eq:augAdjoint}
per integration step.
Even for training on a single time-step, this is more resource intensive
than Flow Matching, which only requires the derivative of the vector field $\frac{\partial v_\theta(x_t, t)}{\partial \theta}$.
Thus, although the memory demands do not increase with additional integration steps of the ODE, the memory required for Path Gradients is greater than with Flow Matching.
Since the computational load associated with calculating the terms varies across different architectures, we discuss them individually for every experiment.\\
For RK4 with 15 integration steps, the ODEs \ref{eq:NODE}, \ref{eq:Divergence} and \ref{eq:augAdjoint} are evaluated in the backward direction ($t: 1 \rightarrow 0$) and the ODE \ref{eq:NODE} and the adjoint method in the forward direction,
amounting in 300 function evaluations, compared to two (the forward and backward call) with Flow Matching.
Thus we expect Flow Matching to be faster than Path Gradients by a factor of at least 150. 

\section{Experiments}
\label{sec:Experiments}

Motivated by the improvements on the toy model, we now turn to more challenging problems, namely the experiments done in \citet{klein2023equivariant} and \citet{klein2024transferable}.
To this end, we use the same architecture, a CNF with an EGNN (E(n) Equivariant Graph Neural Network) \citep{satorras2021en, satorras2021n} for the vector field. % with different atom embeddings .
The architecture in \citet{klein2024transferable} is mostly the same as in \citet{klein2023equivariant}, with the main difference of the encoding of the atoms. While the model in \citet{klein2023equivariant} treats the same atom type as indistinguishable, the model in \citet{klein2024transferable} encodes nearly all atoms differently. Only hydrogens bound to the same atom are treated as indistinguishable. 
For more details on the model architecture see \cref{app:architecture}.
While we maintain the architecture and the initial training process with Flow Matching, we change the later training procedure to include Path Gradients. 
As in \citet{klein2023equivariant}, we investigate optimal transport Flow Matching (OT FM), -- and for the second set of experiments on LJ13 -- equivariant optimal transport Flow Matching (EQ OT FM), and naive Flow Matching (standard FM).  \\
We first evaluate the models on two many particle system with pair-wise Lennard-Jones interactions with 13 and 55 particles.
In addition we investigate Alanine dipeptide (AD2) at $T=300K$ both with a classical and a semi-empirical force field (XTB).
In contrast to \citep{klein2023equivariant,klein2024transferable}, we do not bias the training data towards the positive $\varphi$ state.
%, as this violates the assumptions used to derive Path Gradients and led to unexpected behavior during training.
The bias was originally introduced to make learning problem simpler, as the unlikely state is more prominent \citep{klein2023equivariant}.
This, however, skews the evaluation, since the metrics, namely ESS and Negative log likelihood (NLL), are based on the target distribution, not the biased one.
Moreover, it assumes system-specific knowledge of slow-varying degrees of freedom, information that is typically unavailable for unknown systems.
Finally, we also investigate the 2AA dataset consisting of dipeptides simulated at $T=310K$ with a classical force field \citep{klein2023timewarp}, to evaluate the influence of PG in a transferability. 
For more details on the datasets see \cref{app:datasets}.  
We published code for replicating our experiments \footnote{\href{https://github.com/lenz3000/path-grads-after-fm}{github.com/lenz3000/path-grads-after-fm}}.
In total, we investigate three scenarios: 
\begin{itemize}
    \item  First, fine-tuning with Path Gradients compared to fine-tuning with OT Flow Matching given the same limited computational resources in \Cref{sec:FT-lim-resources}.  
    \item Second, we apply Path Gradients on the FM-trained models using unlimited resources to maximize performance after training with Standard FM, OT FM and EQ OT FM as done in \citet{klein2023equivariant}, see \Cref{sec:FT-forever}. In this second case, we additionally examine the ODE integration length to understand how PG fine-tuning influences the model. 
    \item Finally, we test if the path gradient fine-tuning improves results for transferability in the setting by investigating transferable Boltzmann Generators \citep{klein2024transferable} trained and evaluated on dipeptides, see \Cref{sec:FT-tbg}.
\end{itemize}

%Since Path Gradients expect the samples to come from the target density, we did not bias the dataset by reweighting.

\subsection{Fine-tuning using the same limited resources}
\label{sec:FT-lim-resources}

We enforce comparable memory and wall-time constraints to those used for Flow Matching alone by adjusting batch size and employing gradient accumulation. Consequently, our fine-tuning runs with Path Gradients complete in less wall-time and occupy a similar memory footprint. For full training statistics and implementation details, see \cref{app:Finetuning-experiments} and \Cref{app:alanine}.\\
Replicating the experiments in \citet{klein2023equivariant}, we evaluate how path-gradient fine-tuning impacts model performance. 
Table \ref{tab:finetuned-ourselves} presents a comparison between models trained with pure OT Flow Matching and those optimized using the hybrid approach of OT FM pre-training and path-gradient fine-tuning.\\
We observe that fine-tuning with Path Gradients improves the ESS metrics, i.e. ESS$_p$ and ESS$_q$, across most datasets and models, despite their high variance.  
The hybrid approach reliably improves the NLL for all models and datasets, but one.
The only exception is the standard Flow \citep{klein2023equivariant} on AD2 with XTB, where fine-tuning with FM still performs better on average.
Examining the ESS, we find that both approaches fail to adequately fit the target density.
This supports the hypothesis that Path Gradients are only effective for fine-tuning when the flow already provides a reasonably good approximation of the target.
Notably, on LJ55, our hybrid approach nearly triples the ESS, and on AD2 with XTB, it doubles it \textit{-- in the same training time.}

\begin{table}
  \caption{Comparison between fine-tuning with Optimal Transport Flow Matching and Path Gradients.
  First all models were pre-trained with Optimal Transport Flow Matching like in \citep{klein2023equivariant}. We compare fine-tuning with Flow Matching and Path Gradients.
  For all experiments, we limited the VRAM and runtime to be roughly equivalent. Mean $\pm$ sterr on three runs. \textbf{bold}: sterrs do not overlap.
%\textcolor{red}{The likelihoods between TBG and BG are maybe not comparable, in a similar vein to \citep{kirichenko2020normalizing}. What do you think?}
}
  \label{tab:finetuned-ourselves}
  \centering
  \begin{tabular}{llccc}
    \toprule
    \textbf{Model} & \textbf{Training type} &  \textbf{NLL $(\downarrow)$} & \textbf{ESS$_q$ $\% (\uparrow) $} & \textbf{ESS$_p$ $\% (\uparrow)   $}\\ % & \textbf{Path length $(\downarrow)$} \\
    \cmidrule{1-5}
        & \multicolumn{4}{c}{LJ13} \\
    \cmidrule{2-5}    
    \multirow{2}{*}{\citep{klein2023equivariant}} & Only FM   &$-16.09\pm0.03$ & $54.36 \pm 5.43 $ & $58.18 \pm 0.71 $ \\ % & $2.84\pm0.01$\\
    & Hybrid (ours) & $\mathbf{-16.21 \pm 0.00}$ & $\mathbf{82.97 \pm 0.40 } $ & $\mathbf{82.87 \pm 0.35 }$\\ % & $2.77 \pm 0.01$ \\
    \cmidrule{2-5}
        & \multicolumn{4}{c}{LJ55} \\
    \cmidrule{2-5}    
    \multirow{2}{*}{\citep{klein2023equivariant}} & Only FM   &$-88.45\pm 0.04$ & $3.74\pm 1.06 $ &  $2.97\pm0.08 $ \\ % & $7.53\pm0.02$\\
    & Hybrid (ours) & $\mathbf{ -89.19 \pm 0.05 }$ & $\mathbf{ 11.04 \pm 3.98  }$ & $\mathbf{13.71 \pm 3.15  }$ \\ %& $7.41 \pm 0.14$ \\
    \cmidrule{2-5}
        & \multicolumn{4}{c}{Alanine dipeptide - XTB} \\
    \cmidrule{2-5}
    \multirow{2}{*}{\citep{klein2023equivariant}} & Only FM & $-107.89\pm 0.07 $&$0.74\pm 0.33$ & $0.01 \pm 0.01$ \\ %
    & Hybrid (ours) & $-107.77 \pm 0.18$ & $0.25 \pm 0.37$ & $0.01 \pm 0.01$  \\
    %& Hybrid (ours) 1024 & $-107.83 \pm 0.09$ & $1.04 \pm 0.34$ & $0.01 \pm 0.01$  \\
    \cdashline{2-5}
    \rule{0pt}{2ex}
    \hspace{-1ex}
    \multirow{2}{*}{\citep{klein2024transferable}} & Only FM & $-125.75 \pm 0.01$ & $3.38 \pm 0.50$ & $2.65 \pm 0.50$ \\
    & Hybrid (ours) & $\mathbf{-125.82 \pm 0.02 }$ & $\mathbf{ 7.30  \pm 1.28 }$ & ${6.57 \pm 2.98  } $  \\
    %& Hybrid (ours) (1024) & $\mathbf{-125.86 \pm 0.02 }$ & ${ 4.48  \pm 1.67 }$ & $\mathbf{6.93 \pm 1.09  } $  \\
    \cmidrule{2-5}
        & \multicolumn{4}{c}{Alanine dipeptide - Classical} \\
    \cmidrule{2-5}
    \multirow{2}{*}{\citep{klein2023equivariant}} & Only FM & $-110.14\pm0.01$&$4.88\pm 0.42$ & $0.26 \pm 0.19$ \\
    & Hybrid (ours) & $\mathbf{ -110.30 \pm 0.13 } $ & ${ 2.86 \pm 3.05 }$ & $0.04 \pm 0.06$  \\
    %& Hybrid (ours) (1024) & $\mathbf{ -110.37 \pm 0.05 } $ & $\mathbf{ 8.74 \pm 0.63 }$ & $0.05 \pm 0.01$  \\
    %OT FM full$ +$ Path Grads & $-110.49 \pm 0.10$ & $8.36 \pm 1.49$  \\
    \cdashline{2-5}
    \rule{0pt}{2ex}    
    \multirow{2}{*}{\citep{klein2024transferable}} & Only FM & $-128.01 \pm 0.01$ & $14.42 \pm 2.44$ & $12.21 \pm 1.06$ \\
    & Hybrid (ours) & $\mathbf{-128.26 \pm 0.02 }$ & $\mathbf{ 24.39  \pm 6.86 }$ & ${13.89 \pm 4.91  } $  \\
    %& Hybrid (ours) (1024) & $\mathbf{-128.29 \pm 0.02 }$ & ${ 21.06  \pm 10.25 }$ & $\mathbf{18.21 \pm 3.58  } $  \\
    \bottomrule
  \end{tabular}
\end{table}

\begin{figure}
    \centering
    \includegraphics[width=1\linewidth]{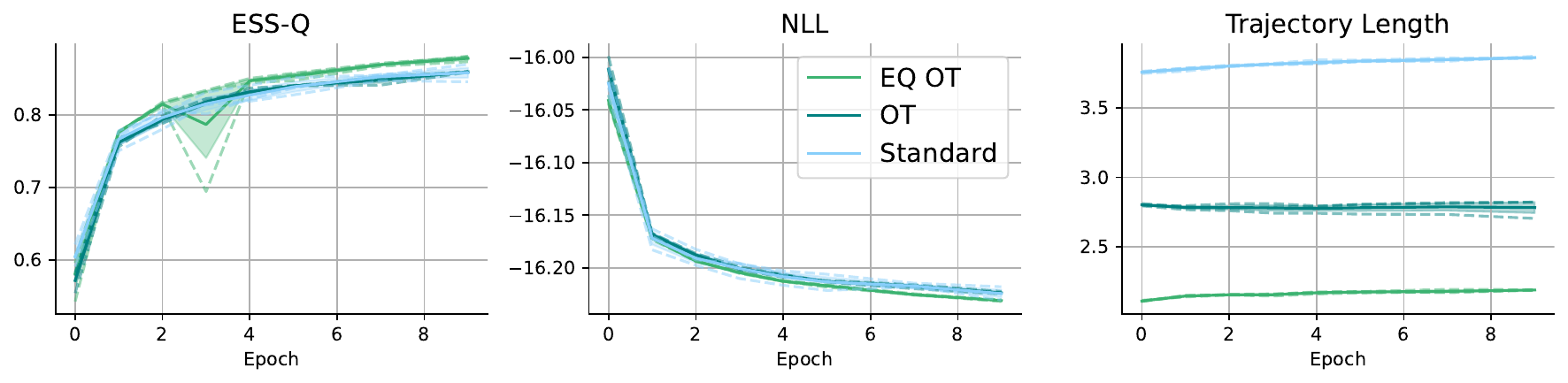}
    \caption{Reverse ESS, NLL and trajectory length for Flows trained with standard FM, Optimal Transport and Equivariant Transport during fine-tuning with Path Gradients on LJ13. We can observe that fine-tuning largely leaves the trajectory length unchanged, while substantially improving performance. Mean $\pm$ sterr over three runs.}
    \label{fig:fine-tune-LJ13}
\end{figure}
\begin{figure}
    \centering
    \includegraphics[width=\linewidth]{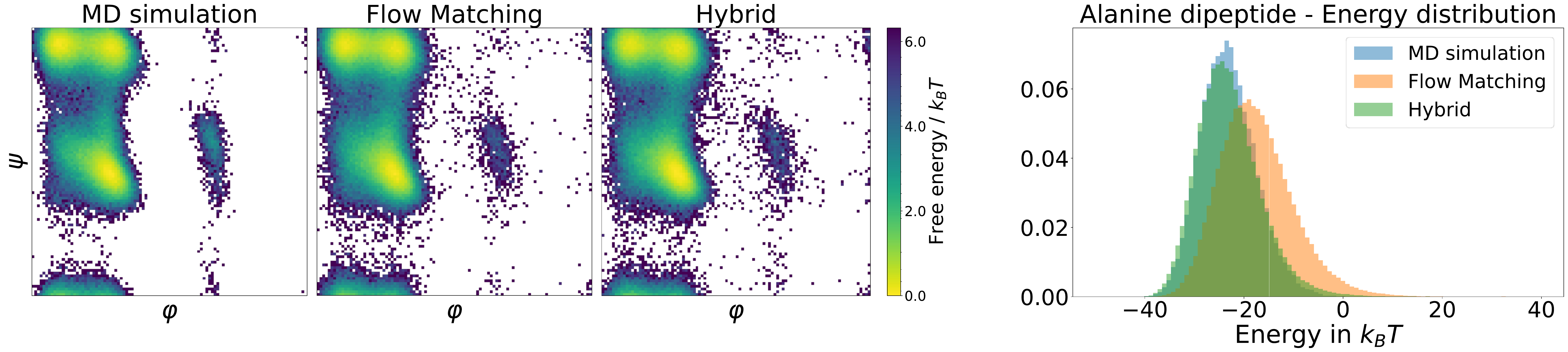}
    \caption{Alanine dipeptide results for the TBG model and the classical force field with and without Path Gradient finetuning. \textbf{Left:} Ramachandran plots for the dihedral angel distribution of a reference MD simulation and non reweighted samples from the different TBG models.  \textbf{Right:} Corresponding energy distributions of generated samples.}
    \label{fig:alanine}
\end{figure}

\subsection{Effect of Path Gradient fine-tuning on flow trajectory length}
\label{sec:FT-forever}
We use the provided saved models from \citep{klein2023equivariant} for LJ13 and investigate the performance and flowed trajectory length during fine-tuning with Path Gradients.
We used a batch-size of 256 and trained with Path Gradients for 10 epochs. In \Cref{fig:fine-tune-LJ13},
we can see that the different pre-trained models have similar NLLs and ESS, but differ in trajectory lengths.
Fine-tuning reliably improves the performance of the flows with relative little change to the trajectory lengths.
For the full statistics see  \Cref{sec:App-Finetune-PathLength}. 
We observe a similar pattern in the other systems investigated. For alanine dipeptide, for instance, generated samples show significantly improved bond-length and energy distributions, yet the global conformational landscape, captured by the $\varphi$ and $\psi$ dihedral-angle distributions in the Ramachandran plot, remains mostly unchanged (see \Cref{fig:alanine} and \Cref{fig:alanine-appendix}).

These experiments show that our proposed hybrid approach is perfectly suited for maximizing performance while keeping the properties of the model.

\subsection{Transferability on dipeptides}
\label{sec:FT-tbg}
Finally, we also investigate whether fine-tuning with Path Gradients improves performance in the transferable setting.
To this end we fine-tune Transferable Boltzmann Generators (TBG) \citep{klein2024transferable}, which were trained on 200 different dipeptides and evaluate on 16 unseen ones, like in \citet{klein2024transferable}.
The experiments show that fine-tuning with path gradients improves the NLL and energies for all evaluated unseen test dipeptides. On average, we observe a relative improvement of around 23\% in the ESS$_q$, reaching an efficiency of $9.79 \%$.
For more details see \cref{sec:dipeptides}.

\section{Discussion}
\label{sec:Conclusion}

We have presented a hybrid training strategy for Boltzmann Generators that uses Flow Matching and Path Gradients. 
We have shown how to efficiently compute Path Gradients with constant memory and without the need for additional samples.
While substantially slower per training step, Path Gradients are a powerful tool for fine-tuning, leveraging first-order information from the energy function -- an underexplored avenue in the scientific machine learning community.
Our results demonstrate that Path Gradients can significantly improve sample quality on the same computational budget as Flow Matching, when the model is already reasonably close to the target distribution.
Our experiments show that Path Gradients only apply minor changes to the flow trajectory and to the variational distribution $q_\theta$ while still substantially increasing the sampling efficiency and performance.

\subsection{Limitations}
\label{sec:Limitations}
Path Gradients come with several limitations.
First, they require access to a well-defined and differentiable energy function, which restricts their use to domains like molecular modeling and excludes standard tasks such as natural image generation.
Second, they rely on unbiased training samples.
Finally, the method is computationally more expensive and tends to improve performance only when the model is already close to the target distribution.
While Path Gradients do not directly speed up the expensive CNF sampling process, they increase the ESS, thereby reducing the total number of samples needed and, hence, at least partially alleviating the high sampling cost.

%{In their empirical investigation about the role of gradient estimators and batch-size,
%\citet{agrawal2024disentangling} concluded
%that Path Gradients are beneficial for training,
%with increased batch-size having an even stronger impact.
%In our context, where we only have access to a limited number of training samples, we can not arbitrarily increase the batch-size, but we can use sample efficient Path Gradients.}

\subsection{Future Work}
Given the similarities between Flow Matching and Diffusion models, extending Path Gradients to diffusion-based frameworks presents an exciting and promising direction.
The dynamics used in our experiments have also been used for molecular conformation generation using Diffusion models \citep{hoogeboom2022equivariant}.
Adoption to their applications would be an interesting and suitable candidate for future work. \\
Furthermore, in model distillation (e.g. \citep{salimans2022progressive}) the gradient information of the larger model is available, even though there might be no first order information of the original data.
Here our approach could help to improve the distillation process.

\subsection{Broader Impact}\label{sec:broader}
This foundational research has no immediate societal impact, but if scalable, it could be used to accelerate drug and materials discovery by replacing MD simulations. Potential risks include misuse for biothreat development and the lack of convergence guarantees, which may lead to incomplete sampling and misleading conclusions.

\section*{Acknowledgements}
LV wants to thank Pan Kessel and Shinichi Nakajima for the helpful discussions. LK gratefully acknowledges support by BIFOLD - Berlin Institute for the Foundations of Learning and Data.

\clearpage
\newpage
\bibliography{references}
\bibliographystyle{icml2025}

%%%%%%%%%%%%%%%%%%%%%%%%%%%%%%%%%%%%%%%%%%%%%%%%%%%%%%%%%%%%%%%%%%%%%%%%%%%%%%%
%%%%%%%%%%%%%%%%%%%%%%%%%%%%%%%%%%%%%%%%%%%%%%%%%%%%%%%%%%%%%%%%%%%%%%%%%%%%%%%
% APPENDIX
%%%%%%%%%%%%%%%%%%%%%%%%%%%%%%%%%%%%%%%%%%%%%%%%%%%%%%%%%%%%%%%%%%%%%%%%%%%%%%%
%%%%%%%%%%%%%%%%%%%%%%%%%%%%%%%%%%%%%%%%%%%%%%%%%%%%%%%%%%%%%%%%%%%%%%%%%%%%%%%
\newpage
\appendix
\onecolumn
\clearpage
\newpage
\section{Appendix}\label{app:A}

\subsection{2D GMM experiment}
The 2D GMM is a Gaussian Mixture model with 4 equally weighted Gaussians 
\begin{equation}
\mathcal{N} \left( 
\begin{bmatrix} a_i \\ b_i \end{bmatrix}, 
\begin{bmatrix} c_i + 0.01 & 0 \\ 0 & d_i + 0.01 \end{bmatrix} 
\right) \, ,
\end{equation}
 with $a_i,b_i,c_i,d_i \sim \mathcal N(0,1)$, $i \in \{1,2,3,4\}$.
The training set consists of 2000 samples, the $KL(p|q_\theta)$ and MSE loss are evaluated on 2048 samples.
We used Adam with default parameters and lr=1e-2 for pure FM/PG and pre-training with FM and lr=5e-3 for finetuning with PG.
All experiments are run on an Intel i7-1165G7 CPU and ignoring the validation, training completes in roughly 45 seconds.

\begin{figure}
    \centering
    \includegraphics[width=0.8\linewidth]{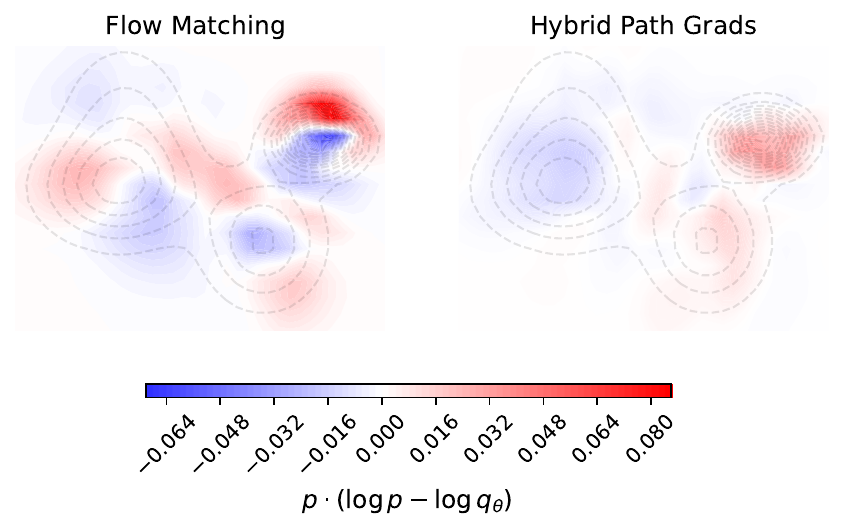}
    \caption{Loss in space after training as done in \cref{fig:2d-GMM}}
    \label{fig:KLparts-2dGMM}
\end{figure}

\cref{fig:KLparts-2dGMM} shows the weighted difference 
\begin{equation}
    p(x) \cdot \log \frac{p(x)}{q_\theta(x)}
\end{equation}
after training.

\subsection{Architecture}\label{app:architecture}
We use the same model architecture as introduced in \citet{klein2023equivariant, klein2024transferable}. We here summarize it briefly, closely following the presentation in \citet{klein2024transferable}.

The underlying normalizing flow model for the Boltzmann Generator is a CNF. The corresponding vector field $v_{\theta}(t,x)$ is parametrized by an $O(D)$- and $S(N)$-equivariant graph neural network (EGNN) \cite{satorras2021n, satorras2021en}.
The vector field $v_{\theta}(x, t)$ consists of $L$ consecutive EGNN layers. The position of the $i$-th particle $x_i$ is updated according to the following set of equations:
%\todo{add tanh}
\begin{align}
    h_i^0 &= (t, a_i), \quad m_{ij}^l=\phi_e\left(h_i^l, h_j^l, d_{ij}^2 \right),\\
    x_i^{l+1}&=x_i^l+\sum_{j\neq i} \frac{\left(x_i^l-x_j^l \right)}{d_{ij}+1}\phi_d(m_{ij}^l),\\
    h_i^{l+1}&=\phi_h\left(h_i^l, m_i^l\right), \quad  m_i^l=\sum_{j\neq i} \phi_m(m_{ij}^l) m_{ij}^l,\\
    v_{\theta}(x^0, t)_i &= x_i^L - x_i^0 - \frac{1}{N}\sum_{j}^N (x_j^L - x_j^0),
\end{align}
where the $\phi_\alpha$ represent different neural networks, $d_{ij}$ is the Euclidean distance between particle $i$ and $j$, $t$ is the time, $a_i$ is an embedding for each particle.

The proposed model architecture in \citet{klein2023equivariant} uses for alanine dipeptide distinct encodings $a_i$ for all backbone atoms and the atom types for all other atoms. 
In contrast, the model architecture in \citet{klein2024transferable} uses distinct encodings for all atoms, except for Hydrogens bond to the same Carbon atom. We refer to this model as \textit{TBG}, which stands for transferable Boltzmann Generator, even though we do not deploy it in a transferable way in this work.

For more details see \citet{klein2023equivariant, klein2024transferable}.

\subsection{Datasets}\label{app:datasets}
Here we provide more details on the investigated datasets. 

We use the same training and test splits as defined in \citet{klein2023equivariant, klein2024transferable}. 
\paragraph{Lennard-Jones systems}
The energy $U(x)$ for the Lennard-Jones systems is given by
\begin{equation}
    U(x) = \frac{1}{2} \left[ \sum_{i,j} \left( \left(\frac{1}{d_{ij}}\right)^{12} - 2 \left(\frac{1}{d_{ij}}\right)^{6} \right)\right],
\end{equation}
where $d_{ij}$ is the distance between particle $i$ and $j$. The authors of \citet{klein2023equivariant} (CC BY 4.0) made the datasets available here: \url{https://osf.io/srqg7/?view_only=28deeba0845546fb96d1b2f355db0da5}. \\
For computing the metrics we use the following number of test samples: \\
LJ13: $ESS_q$: $5 \times 10^5$; $ESS_p$, NLL: $5 \times 10^5$. \\
LJ55: $ESS_q$: $1 \times 10^5$; $ESS_p$, NLL: $1 \times 10^5$.

\paragraph{Alanine dipeptide}
The classical alanine dipeptide dataset was generated with an MD simulation, using the classical \textit{Amber ff99SBildn} force-field at $300\textrm{K}$ for implicit solvent for a duration of $1$ ms \cite{kohler2021smooth} with the \texttt{openMM} library. 
The datasets is available as part of the public \texttt{bgmol}  (MIT licence) repository here: \url{https://github.com/noegroup/bgmol}.  

The alanine dipeptide dataset with the semi empirical force-field, was generated by relaxing $10^5$ randomly selected states from the classical MD simulation. The relaxation was performed with the semi-empirical \textit{GFN2-xTB} force-field for $100$ fs each, using  \citep{xtb} and the ASE library \citep{ase-paper} with a friction constant of $0.5$ a.u. The test set was created it in the same way. 
The authors of \citet{klein2023equivariant} made the relaxed alanine dipeptide with the \textit{semi empirical} force field available here (CC BY 4.0): \url{https://osf.io/srqg7/?view_only=28deeba0845546fb96d1b2f355db0da5}.  

For AD2-XTB, we precompute the target forces of the samples, to re-use them during PG training.\\
For computing the metrics we use the following number of test samples:\\ $ESS_q$: $2 \times 10^5$; $ESS_p$, NLL: $1 \times 10^5$.

\paragraph{Dipetide dataset (2AA)}
The dipeptide dataset was introduced by \citet{klein2023timewarp} and is available at \url{https://huggingface.co/datasets/microsoft/timewarp}. The dataset consists of classical MD trajectories of dipeptides at $310$K. There are $200$ trajectories in the train set, simulated for $50$ns each and $100$ each in the validation and test set, simulated for $1\mu$s each. 

\subsection{Finetuning experiments}
\label{app:Finetuning-experiments}
We used Adam with a learning rate of 1e-4 for PG and only the training set provided.
For each of the training points, we require access to the force of the target.
We use Hutchinson's estimator for estimating the trace of the Jacobian.
Importantly, in order to keep the memory constant, we avoid saving checkpoints.
For our experiments we used A100 GPUs.

For LJ13, we used a batch-size of 64 instead of 256, which uses about 90\% of the original memory (FM $1.52 GB$, PG $1.38GB$).
For fine-tuning we used 2 epochs for Path Gradients, which took 117 min, instead of 135 min for 1000 epochs with Flow Matching.
For LJ55, we use the same batch-sizes resulting in $13.24 GB $ for Flow Matching and $14.76 GB$ for Path Gradients.
We fine-tuned with Path Gradients for 1 epoch, taking 394 minutes instead of 440 min for 400 epochs with Flow Matching.

Because we are measuring the ESS and neg-loglikelihood on the target density, we removed the reweighting from the Alanine Dipeptide data.
If we changed the target density to its reweighted version and computed the gradients of the weighting w.r.t. the samples, we could also straightforwardly apply Path Gradients to the reweighted distribution.
Further, fine-tuning on AD2 with Path Gradients showed some instabilities during training.
To fight these, we employed gradient clipping to a norm of 1 and gradient accumulation to a batch-size of around 1000 in batches of 50. 
The resulting fine-tuning with Path Gradients uses $2.44 GB$ VRAM and 
131 min, compared to 200 minutes with $2.95GB$ with Flow Matching and batch-size 256. 

We used the same number of samples like \citep{klein2023equivariant} for the ESS, but note that a higher number might have been beneficial for more reliable estimates.

For the experiments, we did not run exhaustive hyperparameter tuning.
The Batch-sizes were set to have a similar memory footprint. 
The hybrid PG learning rate was set to the middle between the initial and FM-fine-tuning one after some preliminary experiments.

\subsection{Additional results for alanine dipeptide}\label{app:alanine}
\begin{figure}
    \centering
    \includegraphics[width=\linewidth]{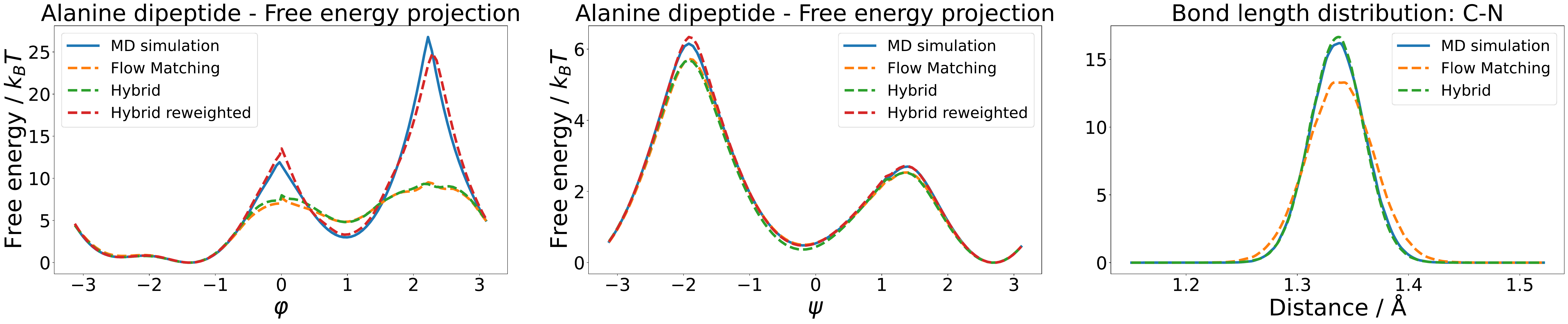}
    \caption{Alanine dipeptide results for the TBG model and the classical force field with and without Path Gradient finetuning. \textbf{Left and middle:} Free energy projection for the $\varphi$ and $\psi$ dihedral angles, respectively.  \textbf{Right:} Bond-length distribution for a Carbon - Nitrogen bond.}
    \label{fig:alanine-appendix}
\end{figure}

Here we present additional alanine dipeptide results for the TBG model in \cref{fig:alanine-appendix}. We observe that the  $\varphi$  and $\psi$ dihedral-angle projections change little after path-gradient fine-tuning. Moreover, reweighting to the target distribution via \cref{eq:importance} succeeds in both cases, though it is more efficient post–fine-tuning, as reflected by a higher ESS (see \cref{tab:finetuned-ourselves}). In contrast, all bond-length distributions align much more closely with the target after fine-tuning, exemplified here by a Carbon - Nitrogen bond in \cref{fig:alanine-appendix}.

\subsection{Additional results for alanine dipeptide - classical with more memory}

For classical AD2, we further investigated the performance of fine-tuning with PGs when allowed a larger memory footprint.
We fine-tuned with batch-size 1024 for two epochs, which resulted in $22GB$ VRAM usage and a runtime of 92 minutes.
In \cref{tab:finetuned-ad2-classical-unlim} we can see that this yielded a further improvement over \cref{tab:finetuned-ourselves}.

\begin{table}[ht]
  \caption{Additional experiment for fine-tuning TBG \citep{klein2024transferable} with Path Gradients on classical Alanine Dipeptide.
  First all models were pre-trained with Optimal Transport Flow Matching like in \citep{klein2023equivariant}. We compare fine-tuning with Flow Matching and Path Gradients.
  Here we did not limit the memory usage and trained PG with batch-size 1024 for 2 epochs. mean $\pm$ sterr over 3 runs.
%\textcolor{red}{The likelihoods between TBG and BG are maybe not comparable, in a similar vein to \citep{kirichenko2020normalizing}. What do you think?}
}
  \label{tab:finetuned-ad2-classical-unlim}
  \centering
  \begin{tabular}{lccc}
    \toprule
     \textbf{Training type} &  \textbf{NLL $(\downarrow)$} & \textbf{ESS$_q$ $\% (\uparrow) $} & \textbf{ESS$_p$ $\% (\uparrow)   $}\\ % & \textbf{Path length $(\downarrow)$} \\
     \midrule
        %& \multicolumn{4}{c}{Alanine dipeptide - Classical} \\
    %\cmidrule{2-5}
     Only FM & $-128.01 \pm 0.01$ & $14.42 \pm 2.44$ & $12.21 \pm 1.06$ \\
    %& Hybrid (ours) limited & $-128.26 \pm 0.02 $ & ${ 24.39  \pm 6.86 }$ & ${13.89 \pm 4.91  } $  \\
     Hybrid (ours) unlimited  & $\mathbf{-128.33 \pm 0.04 }$ & $\mathbf{ 29.47  \pm 2.24 }$ & $\mathbf{19.57 \pm 4.79  } $  \\
    %& Hybrid (ours) (1024) & $\mathbf{-128.29 \pm 0.02 }$ & ${ 21.06  \pm 10.25 }$ & $\mathbf{18.21 \pm 3.58  } $  \\
    \bottomrule
  \end{tabular}
\end{table}

\subsection{Additional results for experiments on flow trajectory length}
\label{sec:App-Finetune-PathLength}

\begin{figure}
\centering
\includegraphics[width=\linewidth]{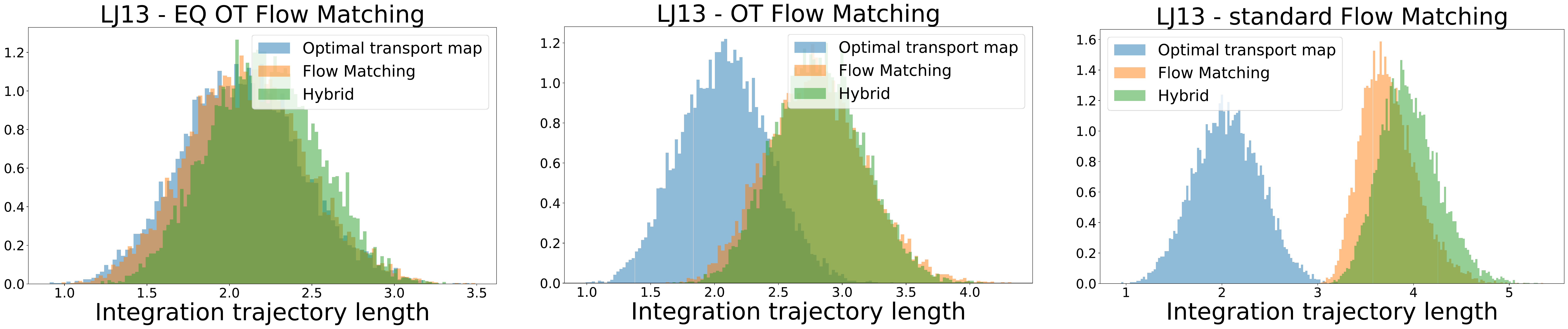}
\caption{Integration trajectory lengths for the Lennard-Jones system with 13 particles. Compared are different Flow Matching models as well as the same models after fine-tuning with Path Gradients (Hybrid).}
\label{fig:pathlength}
\end{figure}

Fine-tuning on the 13-particle Lennard–Jones system used an A100 for 10 epochs (152 min/run) starting from the \citet{klein2023equivariant} checkpoints, with training and evaluation on $5 \times 10^5$ samples. As shown in \cref{tab:performance}, path-gradient fine-tuning boosts NLL and ESS across the board while leaving flow trajectory lengths nearly unchanged (see \cref{fig:pathlength}).
One of the nine runs diverged during training, was discarded and repeated.

\begin{table}[ht]
\centering
\caption{Performance metrics before and after fine-tuning for different methods for the LJ13 system. mean $\pm$ sterr over 3 runs.}
\resizebox{\textwidth}{!}{%
\begin{tabular}{llcccc}
\toprule
\textbf{Method} & \textbf{Fine-tuning} & \textbf{Trajectory Length} & \textbf{NLL} & \textbf{ESS$_q$ (\%)} & \textbf{ESS$_p$ (\%)} \\
\midrule
\multirow{2}{*}{Standard} 
  & before & $3.76 \pm 0.00$ & $-16.02 \pm 0.01$ & $60.45 \pm 0.89$ & $40.23 \pm 20.11$ \\
  & after  & $3.86 \pm 0.00$ & $-16.22 \pm 0.00$ & $85.78 \pm 0.69$ & $85.93 \pm 0.57$ \\
\midrule
\multirow{2}{*}{Optimal Transport} 
  & before & $2.80 \pm 0.01$ & $-16.01 \pm 0.01$ & $57.16 \pm 0.88$ & $57.07 \pm 0.54$ \\
  & after  & $2.78 \pm 0.04$ & $-16.22 \pm 0.00$ & $85.91 \pm 0.03$ & $85.78 \pm 0.02$ \\
\midrule
Equivariant Op-
  & before & $2.11 \pm 0.00$ & $-16.04 \pm 0.00$ & $58.11 \pm 1.93$ & $19.77 \pm 12.36$ \\
  timal Transport & after   & $2.19 \pm 0.00$ & $-16.23 \pm 0.00$ & $87.77 \pm 0.24$ & $58.40 \pm 29.20$ \\
\bottomrule
\end{tabular}
}
\label{tab:performance}
\end{table}

\subsection{Comparison with maximum likelihood training}
We here present comparisons with maximum likelihood training as a baseline on LJ13. To this end we compare to \citet{satorras2021n}, which is nearly the same architecture as in \citet{klein2023equivariant}, but the model is trained via maximum likelihood training instead of flow matching. The results are shown in \cref{tab:LJ13-old}.

\begin{table}[h]
\centering
\begin{tabular}{lcc}
\toprule
\textbf{Method} & $\textbf{NLL}$ & $\textbf{ESS}_q$ in $\%$\\
\hline
ML training \citep{satorras2021n} & $-15.83 \pm 0.07$ & $39.78 \pm 6.19$ \\
Only FM  \citep{klein2023equivariant} & $-16.09 \pm 0.03$ & $54.36 \pm 5.43$ \\
Hybrid approach (Ours) & $-16.21 \pm 0.00$ & $82.97 \pm 0.40$ \\
\bottomrule
\end{tabular}
\caption{Results for the Lennard-Jones system with $13$ particles (LJ13).}
\label{tab:LJ13-old}
\end{table}

\subsection{Additional energy histograms}
\label{sec:energy-hists}
\begin{figure}
\centering
\includegraphics[width=\linewidth]{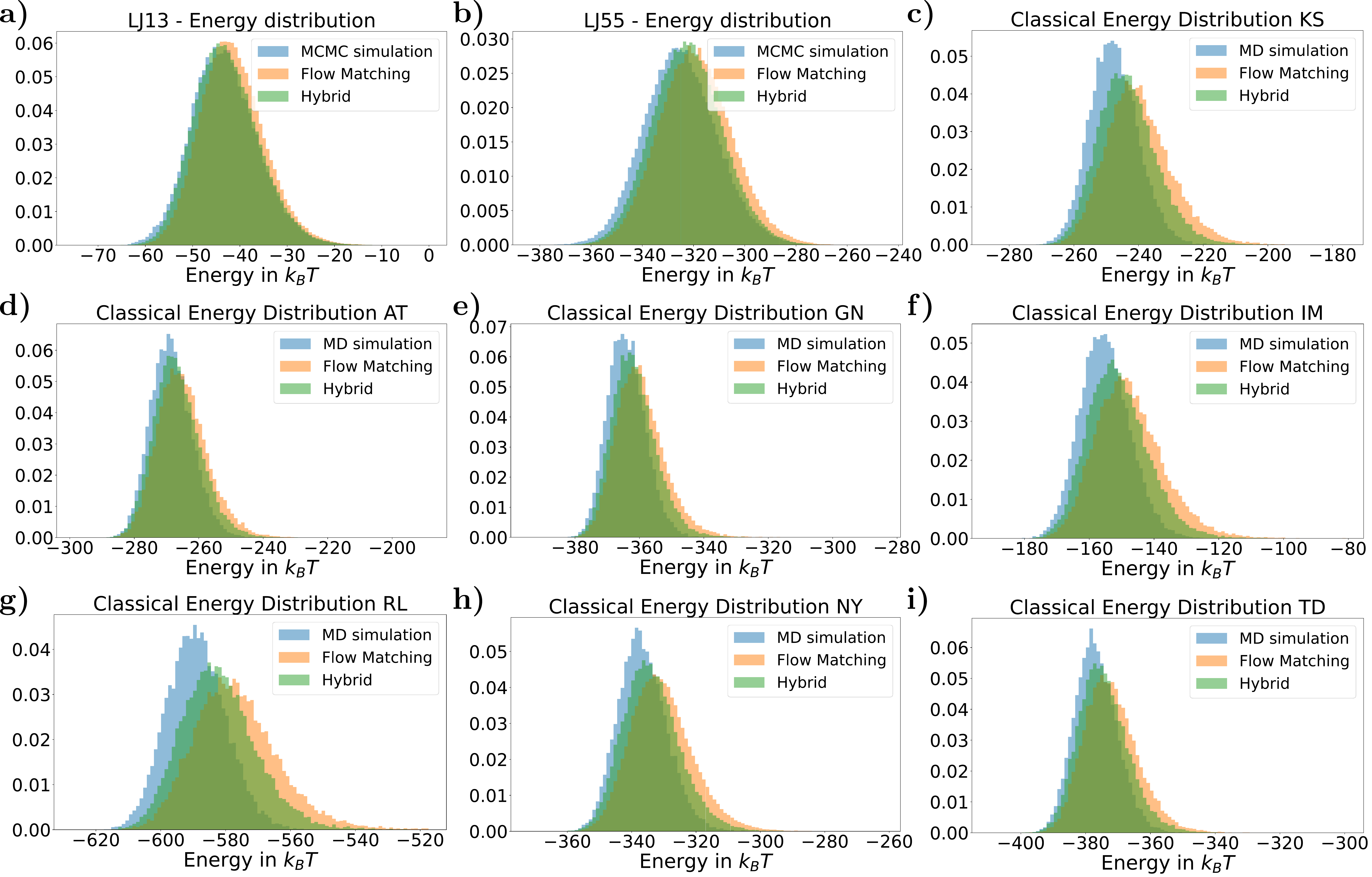}
\caption{Energy histograms for the \textbf{a)} LJ13 systems, \textbf{b)} the LJ55 system, and \textbf{c-i)} Different didpeptides from the testset for the transferable Boltzmann Generator (TBG).}
\label{fig:energy_hists}
\end{figure}

We show additional energy histograms of the systems investigated in \cref{sec:Experiments} in \cref{fig:energy_hists}. The average energy for every investigated system is smaller and closer to the target energy distribution after PG finetuning. 

\subsection{Results for transferable Boltzmann Generators (TBG) evaluated on dipeptides}
\label{sec:dipeptides}
We applied the hybrid approach to transferable Boltzmann Generators (TBG) on dipeptides as introduced in \citet{klein2024transferable}. We again fine-tune the pretrained Boltzmann Generator from \citet{klein2024transferable} with path gradients. Training took 6 days on an A100 with learning rate 0.00001. This transferable Boltzmann Generator is trained on a subset of all possible dipeptides and evaluated on unseen ones (for more details see \cref{app:datasets}). Due to the expensive evaluation, we evaluated the model on $16$ test dipeptides and chose the same subset as in \citet{klein2024transferable}.
Our experiments show that fine-tuning with path gradients improves the NLL and energies for all evaluated unseen test dipeptides with an average improvement to -100.93 from -100.72 NLL and also shows average improvements for the ESS$_q$ of around 23\% to an efficiency of $9.79
\%$  as shown in \cref{tab:dipeptides}. An example energy histogram is shown in \cref{fig:energy_hists}c.

\begin{table}[ht]
\centering
\begin{tabular}{lcccc}
\toprule
\textbf{Dipeptide} & \textbf{NLL (Before)} & \textbf{NLL (After PG)} & \textbf{ESS$_q$ (Before) in $\%$} & \textbf{ESS$_q$ (After PG) in $\%$} \\
\hline
KS & -100.82 & -100.99 & 6.0 & 4.6 \\
AT & -75.49 & -75.61 & 20.8 & 21.5 \\
GN & -65.45 & -65.55 & 19.2 & 25.6 \\
LW & -148.91 & -149.22 & 0.4 & 3.6 \\
NY & -118.24 & -118.47 & 9.5 & 10.5 \\
IM & -106.80 & -107.01 & 3.1 & 4.6 \\
TD & -81.09 & -81.19 & 4.1 & 11.6 \\
HT & -103.06 & -103.27 & 0.2 & 5.5 \\
KG & -88.60 & -88.79 & 5.0 & 7.4 \\
NF & -116.45 & -116.67 & 3.7 & 12.2 \\
RL & -135.71 & -136.12 & 1.3 & 1.5 \\
ET & -89.94 & -90.08 & 6.3 & 4.1 \\
AC & -63.90 & -63.91 & 31.8 & 33.3 \\
GP & -71.63 & -71.77 & 10.5 & 8.0 \\
KQ & -119.36 & -119.68 & 4.3 & 2.0 \\
RV & -126.28 & -126.62 & 1.3 & 0.6 \\
\hline
Average & -100.73 & -100.93 & 7.96 & 9.79 \\
\bottomrule
\end{tabular}
\caption{Comparison of NLL and ESS before and after PG for different dipeptides from the testset.}
\label{tab:dipeptides}
\end{table}

\subsection{About the different estimators}
\label{app:Theory}
In the following, we first discuss the Maximum Likelihood (ML) and Path Gradient (PG) estimators for the forward KL and show benefits of PG over FM for a toy example. 

\subsubsection{The Maximum Likelihood and Path Gradient Estimators}
\label{app:estimators}

In short: both the ML and PG estimators are unbiased and consistent, but they differ in variance. For PG, we can give guarantees about their variance, once $q_\theta$ equals to $p$, see e.g. \citep{Vaitl2024, roeder2017sticking, tucker2018doubly, vaitl2022path}.

Both Maximum Likelihood and the Path Gradient estimators optimize the Forward KL Equation \ref{eq:InitialKLForward}. Let’s look at both estimators in detail and compare them. A full derivation can be found in Appendix B.3.2 of \citet{vaitl2023fast} \& Chapter 4 of \citet{Vaitl2024}.

To obtain the estimator $\mathcal G_{ML}$, we first observe that the first term in the KL divergence is constant w.r.t.~$\theta$. 
\begin{equation} 
KL(p|q_\theta) = \E_{p(x_1)} \left[ \log p(x_1) - \log q_\theta(x_1) \right] \, ,
\end{equation} 
which means that it does not enter in the gradient if we directly estimate the gradients via an MC estimator.
\begin{align}
\frac{d KL(p|q_\theta)}{d \theta} &= - \E_p(x_1) \left[\frac{d}{d \theta} \log q_\theta(x_1) \right] \nonumber \\
\approx \mathcal G_{ML}  &= -\frac{1}{N} \sum_{i=1}^{N} \frac{d }{d \theta} \log q_\theta(x_1^{(i)}) \, , x_1^{(i)} \sim p \, .
\end{align}

For the exact calculation of the estimator, we decompose it fully with 
\begin{equation}
    q_\theta(x_1) = q_0(T_\theta^{-1}(x_1)) \left| \det \frac{\partial T_\theta^{-1}(x_1)}{\partial x_1} \right|  \, .
\end{equation}

The actual calculation for the ML estimator then is 
\begin{equation}
\mathcal G_{ML} = -\frac 1 N \sum_{i=1}^{N} \frac{d }{d \theta} \left(\log q_0(T_\theta^{-1}(x_1)) + \log \left| \frac{\partial T_\theta^{-1}(x_1)}{\partial x_1}\right| \right) \label{eq:*} \, .
\end{equation}

For PG, we use Eq. \ref{eq:PG-FWKL} to directly obtain the MC estimator $\mathcal G_{PG}$ 
\begin{align}
\frac{d}{d \theta} KL(p_{0, \theta}| q_0) &= E_{x_1 \sim p_1} \left[\frac{\partial }{\partial x_0} \left(\log \frac{p_{0, \theta}}{q_0}(x_0) \right) \frac{\partial T_\theta^{-1}(x_1)}{\partial \theta} \right] \nonumber \\
 \approx \mathcal G_{PG} &= \frac 1 N \sum_{i=1}^N \frac{\partial }{\partial x_0^{(i)} } \left(\log p_{0, \theta}(x_0^{(i)}) - \log q_0(x_0^{(i)}) \right) \frac{\partial T_\theta^{-1}(x_1^{(i)})}{\partial \theta} \, , x_1^{(i)} \sim p \, .
\end{align}
Here $x_0^{(i)}$ is a shorthand for $T_\theta^{-1}(x_1^{(i)})$.
If we again decompose the term we expose the full calculation 
\begin{equation}
\mathcal G_{PG} = \frac 1 N \sum_{i=1}^{N} \frac{\partial }{\partial x_0^{(i)} } \left(\log p(T_\theta(x_0^{(i)})) + \log \left| \frac{\partial T_\theta(x_0)}{\partial x_0}\right| - \log q_0(x_0^{(i)}) \right) \frac{\partial T_\theta^{-1}(x_1^{(i)})}{\partial \theta} \label{eq:**}
\end{equation}

Comparing \eqref{eq:*} and \eqref{eq:**}, we see that the path gradient estimator incorporates $\mathcal G_{PG}$ the gradient information of the target $p$ --  $\frac{\partial \log p(T_\theta(x_0^{(i)})}{\partial x_0^{(i)}}$ -- while $\mathcal G_{ML}$ -- and by construction also $\mathcal G_{FM}$ -- does not.

\subsubsection{Variance of the estimators}

Opposed to $\mathcal G_{ML}$ and $\mathcal G_{FM}$, we have nice guarantees about the variance of the PG estimator $\mathcal G_{PG}$ at the optimum and close to it.

We first recapitulate results from previous work and then, by using a simple example, show that $\mathcal G_{FM}$ does not necessarily exhibit zero variance at the optimum.

In general, the variance of the Path Gradient estimators $\mathcal G_{PG}$ is bounded by the squared Lipschitz constant of the term 
\begin{equation}
\left(\log p_{0, \theta}(x_0) - \log q_0(x_0) \right) = \left(\log p(x_1) - \log q_\theta(x_1) \right) \, ,
\end{equation}
see \citet{mohamed2019monte} Section 5.3.2. \\
Thus, if the target density $p$ is not well approximated by $q_\theta$, the variance of the gradient estimator can be large and training with PG might not be beneficial. If $\log p$ and $\log q_\theta$ are close in the sense that the Lipschitz constant of their difference is small, we can assume path gradient estimators to be helpful.
\paragraph{Gradient estimators at the optimum}

In the case of perfect approximation, i.e. $q_\theta(x) = p(x) \forall x \in \mathcal X$, the following statements about the ML estimator and the Path Gradient estimators are known. The gradients for path gradient estimators are deterministically $0$, i.e. $\E[\mathcal G_{PG}] = 0, \Var[\mathcal G_{PG}] = 0$, while for ML the variance is generically nonzero $\E[\mathcal G_{ML}] = 0, \Var[\mathcal G_{ML}] = \frac 1 N \mathcal I_\theta$. Where $\mathcal I_\theta$ is the Fisher Information Matrix of $p_{0, \theta}$ \citep{vaitl2022gradients}.

\paragraph{Why is that?}
Already the review by \citet{papamakarios2019normalizing} notes in Section 2.3.3 that the duality of the KL divergence means that fitting the model $q_\theta$ to the target $p$ using ML is equivalent to fitting $p_{0,\theta}$ to the base $q_0$ under the reverse KL. 
This means that the results from \citet{Vaitl2024} directly hold. We only adapted the notation.

\subsubsection{Variance of $\mathcal G_{FM}$ for a toy example}

We do not have a term for the variance of $\mathcal G_{FM}$ for general settings, but we can show that it does not deterministically vanish like for $\mathcal G_{PG}$ via a simple example. This means better behavior for PG than for FM at the optimum, which we also verified empirically.
Our assumptions aim to simplify the example as much as possible.

\begin{itemize}
    \item  First, assume the standard loss for Flow Matching.
    \item Further, assume the two densities to be the same $D$-dimensional Normal distribution \\
    $q_0 = p = \mathcal N(0,I)$.
    \item Finally, assume the CNF is the identity parametrized by a single parameter $\theta$, i.e. \\
    $v_\theta = \theta = 0$.
\end{itemize}

In this example $q_\theta(x_1) = q_0(x_1) \cdot I$ approximates the target density $p$ perfectly and the optimal gradient estimator is $0$. 
Yet, in this setting the variance of the estimator $\mathcal G_{FM}$ is non-zero: 
\begin{equation}
\Var[\mathcal G_{FM}] =\frac 8 {ND} \, , 
\end{equation}
preventing the model from staying at the optimum during training.
\paragraph{Proof:}

The gradient estimator for the FM loss Eq \eqref{eq:CFM} is 
\begin{equation}
    \mathcal G_{FM} = \frac 1 N \sum_i \frac{\partial }{\partial \theta} \left(|| v_\theta - ( x_1^{(i)} - {\tilde x}_0^{(i)}) ||^2\right) \, ,
\end{equation}
where ${{\tilde x}_0^{(i)}, x_1^{(i)} \sim \mathcal N(0,I)}$ \footnote{Note that here ${\tilde x}_0^{(i)}$ and $x_1^{(i)}$ are independently sampled. Before $x_0^{(i)}$ was the transformed sample}.

First, we break down the terms 
\begin{align}
    \mathcal G_{FM} &= \frac{1}{N} \sum_{i=1}^{N} \frac{\partial }{\partial \theta} \frac{1}{D} \sum_{d=1}^{D}\left(v_{\theta, d} +\tilde x_{0,d}^{(i)} - x_{1,d}^{(i)} \right)^2 \nonumber \\
    &= \frac{1}{ND} \sum_{i=1}^N \sum_{d=1}^D 2 \left(v_{\theta, d} + \tilde x_{0,d}^{(i)} - x_{1,d}^{(i)}\right) \frac{\partial v_{\theta,d}}{\partial \theta} \, .
\end{align}

Because $v_\theta$ is parametrized by $\theta$, we set in $\frac{\partial v_{\theta,d}}{\partial \theta}=1$ and $v_{\theta,d}=0$ and separate the terms 
\begin{equation}
    \mathcal G_{FM}= \frac {1} {ND} \sum_i \sum_d 2 ( \tilde x_{0,d}^{(i)} - x_{1,d}^{(i)}) = \frac 2 {ND} \left(\sum_d \sum_i \tilde x_{0,d}^{(i)} - \sum_d \sum_i x_{1,d}^{(i)}\right).
\end{equation} 

We can compute the distribution of $\mathcal G_{FM}$ by using the property 
\begin{equation}
    \sum_{i=1}^N a_i \sim \mathcal N(0, N \sigma_a^2) \text{ for } a \sim \mathcal N(0, \sigma_a^2) \, .
\end{equation}
So the sum over dimensions and samples follows the normal distribution with variance $DN$:
\begin{equation}
    \sum_d \sum_i \tilde x_{0,d}^{(i)} \sim N(0, D N)
\end{equation} and the difference has twice its variance 
\begin{equation}
    \sum_d \sum_i \tilde x_{0,d}^{(i)} - \sum_d \sum_i x_{1,d}^{(i)} \sim \mathcal N(0, 2 D N) \, .
\end{equation}

The expectation of the estimator $\mathcal G_{FM}$ then is 0 and the variance is simply
\begin{equation}
    \Var[\mathcal G_{FM}] = \frac 4 {N^2D^2} \Var[(\sum_d \sum_i \tilde x_{0,d}^{(i)} - \sum_d\sum_i x_{1,d}^{(i)})] = \frac 4 {N^2D^2} 2DN = \frac 8 {ND} \, .
\end{equation}
\qed

Interestingly, the derivative $\mathcal G_{FM}$ is invariant to re-ordering because the sums of $x_1$ and $\tilde x_0$ are invariant to permutation. The result thus also holds for OT-FM.
This property is due to the simple assumption, the vector field $v_\theta$ is independent of $x_t$.

\subsection{Pseudocode for forward KL path gradients via augmented adjoint}
\begin{algorithm}[H]
\caption{Augmented Adjoint Dynamics}
\begin{algorithmic}[1]
    \Function{Forward}{$t, x_t, \nabla \log q$}
        \State $\dot{x}, \dot {\text{div}} \gets \text{black\_box\_dynamics}(t, x_t)$
        \State $\dot{\nabla \log q} \gets \text{gradient}_{\text{x}} \left[ -\nabla \log q \cdot \dot{x} - \dot {\text{div}} \right]$
        \State \Return $\dot x, \dot{\nabla \log q}, -\dot{\text{div}}$
    \EndFunction
\end{algorithmic}
\end{algorithm}

\vspace{1em}

\begin{algorithm}[H]
\caption{Pathwise Gradient Estimator}
\begin{algorithmic}[1]
\Function{PathwiseGradientEstimate}{$x_1, \text{prior}, \text{target}, \text{flow}$}
    \State $ \log p_1 \gets  -\text{target.energy}(x_1)$
    \State $\nabla \log p_1 \gets \text{gradient}_{x_1}( \log p_1)$
    \Comment{Integrate using Augmented Adjoint state method}
    \State $x_0, \nabla \log p_{0, \theta}, \log |\det J| \gets \text{flow.integrateAugAdjoint}(x_1, \nabla \log p_1), \texttt{inverse=True})$
    \State $\log q_0 \gets -\text{prior.energy}(x_0)$
    \State $\nabla \log q_0 \gets \text{gradient}_{x_o} (\log q_0)$
    \Comment{Compute gradient of loss w.r.t. sample $x_0$}
    \State $\nabla_{x_0} \mathcal{L} \gets \frac 1 N \left( \nabla \log p_{0, \theta} - \nabla \log q_0 \right)$
    \Comment{Backpropagate using standard Adjoint state method}
    \State $\text{path gradients} \gets \text{gradient}_\theta \left(x_0 \cdot \texttt{detach}(\nabla_{x_0} \mathcal{L})] \right)$ 
\EndFunction
\end{algorithmic}
\end{algorithm}

\subsection{Code libraries}
\label{app:code-libraries}

We primarily use the following code libraries: \textit{PyTorch}  (BSD-3) \citep{NEURIPS2019_9015}, \textit{bgflow} (MIT license) \citep{noe2019boltzmann, kohler2020equivariant}, \textit{torchdyn} (Apache License 2.0) \citep{politorchdyn}. Additionally, we use the code from \citep{satorras2021n} (MIT license) for EGNNs, as well as the code from \citep{klein2023equivariant} (MIT license) and \citep{klein2024transferable} (MIT license) for models and dataset evaluations.

%%%%%%%%%%%%%%%%%%%%%%%%%%%%%%%%%%%%%%%%%%%%%%%%%%%%%%%%%%%%%%%%%%%%%%%%%%%%%%%
%%%%%%%%%%%%%%%%%%%%%%%%%%%%%%%%%%%%%%%%%%%%%%%%%%%%%%%%%%%%%%%%%%%%%%%%%%%%%%%

\end{document}